# A Novel Ontology-guided Attribute Partitioning Ensemble Learning Model for Early Prediction of Cognitive Deficits using Quantitative Structural MRI in Very Preterm Infants

Zhiyuan Li[1,2,4,+], Hailong Li[1,2,5,6,+], Adebayo Braimah[1,2], Jonathan R. Dillman[1,2,3,6], Nehal A. Parikh[5,7], Lili He[1,2,3,5,6,*]

[1]Department of Radiology, Cincinnati Children's Hospital Medical Center, Cincinnati, OH, USA

[2]Imaging Research Center, Cincinnati Children's Hospital Medical Center, Cincinnati, OH, USA

[3]Department of Radiology, University of Cincinnati College of Medicine, Cincinnati, OH, USA

[4]Department of Electronic Engineering and Computer Science, University of Cincinnati, Cincinnati, OH, USA

[5]Center for Prevention of Neurodevelopmental Disorders, Perinatal Institute, Cincinnati Children's Hospital Medical Center, Cincinnati, OH, USA

[6]Artificial Intelligence Imaging Research Center, Cincinnati Children's Hospital Medical Center, Cincinnati, OH, USA

[7]Department of Pediatrics, University of Cincinnati College of Medicine, Cincinnati, OH, USA

+ Equal contribution co-first authors

*Correspondence to:

Lili He, PhD, Cincinnati Children's Hospital Medical Center, 3333 Burnet Avenue, MLC 5033, Cincinnati, OH 45229

E-mail: Lili.He@cchmc.org

# Highlights

- Brain structural MRI features are used to predict cognitive deficits in neonates
- Ontology-guided attribute partitioning method is presented for feature clustering
- Brain ontology knowledge is used to facilitate cognitive deficit prediction
- Ensemble learning integrates maturation and geometric features for the prediction

# Abstract:

Structural magnetic resonance imaging studies have shown that brain anatomical abnormalities are associated with cognitive deficits in preterm infants. Brain maturation and geometric features can be used with machine learning models for predicting later neurodevelopmental deficits. However, traditional machine learning models would suffer from a large feature-to-instance ratio (*i.e.*, a large number of features but a small number of instances/samples). Ensemble learning is a paradigm that strategically generates and integrates a library of machine learning classifiers and has been successfully used on a wide variety of predictive modeling problems to boost model performance. Attribute (*i.e.*, feature) bagging method is the most commonly used feature partitioning scheme, which randomly and repeatedly draws feature subsets from the entire feature set. Although attribute bagging method can effectively reduce feature dimensionality to handle the large feature-to-instance ratio, it lacks consideration of domain knowledge and latent relationship among features. In this study, we proposed a novel Ontology-guided Attribute Partitioning (OAP) method to better draw feature subsets by considering the domain-specific relationship among features. With the better-partitioned feature subsets, we developed an ensemble learning framework, which is referred to as OAP-Ensemble Learning (OAP-EL). We applied the OAP-EL to predict cognitive deficits at 2 years of age using quantitative brain maturation and geometric features obtained at term equivalent age in very preterm infants. We demonstrated that the proposed OAP-EL approach significantly outperformed the peer ensemble learning and traditional machine learning approaches.

# 1. Introduction

The prevalence of neurodevelopmental impairments remains very high for very preterm infants (gestational age; GA ≤32 weeks), though the global infant mortality rate has been reduced to approximately 11.1% [1]. Around 35-40% of very preterm infants develop cognitive deficits at 2 years of corrected age [2, 3]. Cognitive deficits would result in difficulties in academic performance and social abilities, affecting the entire life of those very preterm infants. Unfortunately, an accurate clinical diagnosis of cognitive deficits is currently unavailable for very preterm infants until 3-5 years of age in early childhood, thereby, the absence of prompt treatment leads to missing optimal neuroplasticity period of brain development when interventions can exert the greatest impact on prevention. Thus, a timely and accurate risk stratification approach is desirable to address the need for early prediction of cognitive deficits in very preterm infants.

Multiple structural magnetic resonance imaging (sMRI) studies have shown that several brain anatomical abnormalities are associated with cognitive deficits in preterm infants [4-7]. Altered cortical development has been detected on brain sMRI images in very preterm infants at term-equivalent age. For example, greater cortical thickness in frontal, insular, and anterior parietal cortices was observed in preterm infants compared with term infants [8-10]. These studies demonstrate the promise of brain maturation and geometric features as predictive biomarkers for later neurodevelopmental deficits. Recently, we developed a machine learning model to predict neurodevelopmental outcomes at 2-year corrected age using brain geometric features (*e.g*., volume, cortical thickness, etc.) derived from T2-weighted MRI scans collected at term-equivalent age in 110 very preterm infants [11], demonstrating the predictive abilities of those features for abnormal neurodevelopment. However, our traditional machine learning model still suffered from a large feature-to-instance ratio (*i.e*., a large number of features but a small number of instances/samples).

Ensemble learning is a machine learning paradigm that strategically generates and integrates a library of machine learning classifiers, referred to as base-classifiers. Unlike traditional machine learning models that only learn one hypothesis, ensemble learning defines a set of hypotheses using base-classifiers in the model library and summarizes them into a final decision. Since each base-classifier has its own strengths and weaknesses, it, therefore, is natural to expect that a learning method that takes advantage of multiple bass-classifiers would lead to superior performance beyond the level obtained by any of the individual classifiers [12]. In the last decade, the ensemble learning model has been successfully used on a wide variety of predictive modeling problems to boost model performance [13].

Building a diverse base-classifier library is essential in any ensemble learning strategy. Attribute (*i.e*., feature) bagging (also known as random subspace) method [14, 15] is the most commonly used feature partitioning scheme, which randomly and repeatedly draws feature subsets from the entire feature set to train base-classifiers, instead of using the whole feature set. Attribute bagging method is able to effectively reduce feature dimensionality for each base classifier and to increase model diversity, offering an elegant feature partitioning solution to handle the large feature-to-instance ratio in neuroimaging studies [16]. However, attribute bagging through random drawing lacks consideration of domain knowledge and latent relationship among features. For example, random feature drawing simply treats the attributes of "left amygdala volume" and "right amygdala volume" as two anonymous attributes (*i.e*., only considers their numerical values), without noting that both quantify "volumes"; but one for "left amygdala", and the other for "right amygdala".

Ontology is defined as an explicit specification of a "conceptualization" or "knowledge" in a domain of interest [17-19], and it has been employed for knowledge encoding, sharing, and storing [20-22]. Ontology driven techniques are increasingly being employed in a variety of biomedical research studies, such as protein-protein interactions prediction [23], clinical diagnosis [24], and biological function reasoning [25]. In this study, we proposed an Ontology-guided Attribute Partitioning (OAP) method to better draw feature subsets by considering the domain-specific relationship among features, which are not considered by the standard attribute partitioning methods (*e.g*., attribute bagging method) [14, 15]. With the better partitioned feature subsets, we trained and integrated a stacking/ensemble of diverse individual base-classifiers. We refer to this framework as OAP-Ensemble Learning (OAP-EL). We applied the OAP-EL to predict cognitive deficits at 2 year of age using quantitative brain maturation and geometric features obtained at term equivalent age in very preterm infants. We tested the hypothesis that the proposed OAP-EL approach can significantly outperform the peer ensemble learning approaches with attribute bagging method.

## 2. Materials and Methods

### 2.1. Overview

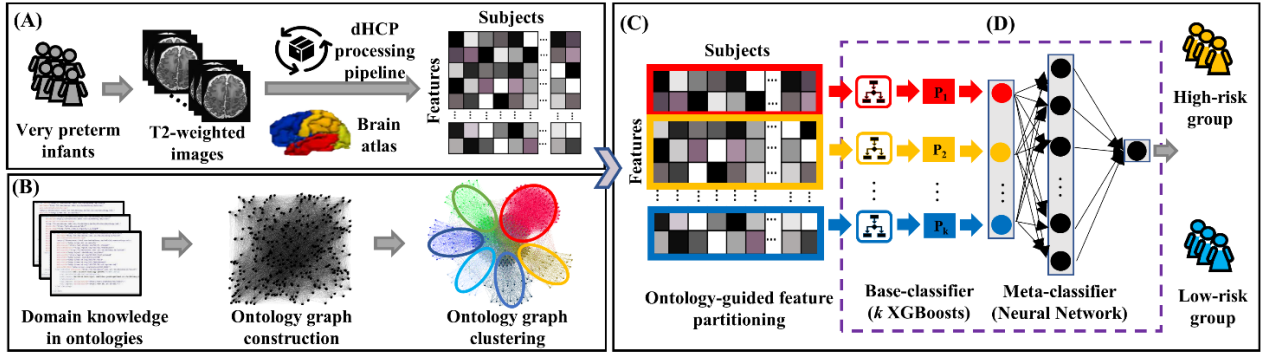

**Figure 1.** Schematic diagram of OAP-EL for early prediction of cognitive deficits at 2 years corrected age using brain maturation and geometric features derived from T2-weighted MRI acquired at term-equivalent age in very preterm infants. (A) Brain maturation and geometric feature extraction; (B) Ontology graph construction and clustering; (C) Ontology-guided feature partitioning; and (D) Base-classifiers training and ensembling.

Our clinical task in this study is to stratify the risk of cognitive deficits in very preterm infants at 2 years corrected age based on quantitative brain maturation and geometric features acquired on structural MRI at term-equivalent age. As shown in **Figure 1**, we first extract hundreds of brain maturation and geometric features from T2-weighted MRI data acquired at term-equivalent age for individual very preterm infants using the Developing Human Connectome Project (dHCP) processing pipeline [26] **(Figure 1A)**. Next, based on two prior defined ontologies, which respectively describe brain parcellation (*e.g.*, frontal, temporal, parietal, etc.) [27], and brain geometry and maturation (cortical thickness, sulcal depth, curvature, cortical surface area, etc.) [26], we construct an unweighted ontology graph, in which brain maturation and geometric features are considered as vertices and ontology-derived relationships are edges. We then conduct ontology graph clustering **(Figure 1B)** to partition brain maturation and geometric features into $k$ non-overlapping feature subsets **(Figure 1C)**. With $k$ feature subsets, we train $k$ base-classifiers (i.e., eXtreme Gradient Boosting (XGBoost) classifiers [28] in this work). Finally, a neural network is used as the meta-classifier to integrate $k$ individual base-classifiers for risk stratification **(Figure 1D)**.

## 2.2. MRI Data Acquisition and Follow-up Cognitive Assessment

The cohort from the CCHMC site consisted of very preterm infants from the Cincinnati Infant Neurodevelopment Early Prediction Study (referred to as CINEPS Cohort) [29]. All infants born at or before 32 weeks of GA, between September 2016 and November 2019, who were cared for in one of the five level III/IV Cincinnati, Ohio Neonatal Intensive Care Units (NICUs), including CCHMC, University of Cincinnati Medical Center, Good Samaritan Hospital, St. Elizabeth's Healthcare, and Kettering Memorial Hospital, were eligible for CINEPS inclusion. The other cohort from the NCH site (referred to as the Columbus Early Prediction Study [COEPS]) enrolled very preterm infants born at 31 weeks GA or younger (between December 2014 and April 2016) from one of the four level III/IV Columbus, Ohio NICUs, including NCH, Ohio State University Medical Center, Riverside Hospital, and Mount Carmel St. Ann's Hospital. After initial recruitment discussion and related consent, all very preterm infants were medically transported to either CCHMC site (CINEPS cohort) or NCH site (COEPS cohort) for their MRI scans after NICU discharge. Infants were excluded from both CINEPS and COEPS cohort, if any of known congenital brain anomalies or severe injury was met [30]. The CINEPS cohort was used for model development and internal cross validation, while the independent COEPS cohort was used as an unseen testing dataset for external validation.

At CCHMC site, CINEPS subjects were imaged at 39-44 weeks postmenstrual age (PMA) during unsedated sleep on a 3T Philips Ingenia scanner with a 32-channel receiver head coil. Acquisition parameters for axial T2-weighted turbo spin-echo sequence are repetition time (TR) = 8300 ms, echo time (TE) = 166 ms, FA = 90°, resolution $1.0 \times 1.0 \times 1.0$ mm$^3$, and time 3:53 min. At NCH site, COEPS subjects were scanned at 38-43 weeks PMA during unsedated sleep on a 3T MRI scanner (Skyra; Siemens Healthcare) using a 32-channel receiver head coil. Acquisition parameters for axial T2-weighted fast spin-echo sequence are TR = 9500 ms, TE = 147 ms, FA = 150°, resolution $0.93 \times 0.93 \times 1.0$ mm$^3$, and time 4:09 min.

All subjects were assessed at 2 years corrected age using the well-established Bayley Scales of Infant and Toddler Development III (Bayley III) test [31]. The Bayley III Cognitive subtest scores (on a scale of 40 to 160, with a mean of 100 and standard deviation shown (SD) of 15) served as the primary measures of the infant's cognitive development functioning level.

## 2.3. MRI Data Preprocessing and Brain Maturation and Geometric Feature Extraction

We preprocessed T2-weighted MRI data of each subject and extracted brain maturation and geometric features using dHCP structural pipeline [26]. Briefly, the pipeline conducted bias field correction [32], brain extraction [33, 34], and brain surface reconstruction [35]. The pipeline segmented the whole brain image into 87 region-of-interests (ROIs) based on an age-matched neonatal volumetric atlas [27, 33]. For individual brain ROIs, up to six different types of brain maturation and geometric metrics were calculated, including volume, thickness, sulcal depth, curvature, gyrification index, and surface area. It is worth noting that, for some ROIs, not all aforementioned metrics are available. For example, dHCP pipeline only extracts the volume metric for the hippocampus. In addition, dHCP pipeline calculates both absolute and relative measures for volume and surface area metrics, which are highly dependent on individual brain sizes. The relative measures are calculated using the absolute measures divided by whole-brain measures. A total of 510 brain maturation and geometric features were calculated by the dHCP pipeline from the whole brain. For both volume and surface area metric types, we only retained relative measures to reduce the impact of individual brain size variance. In this way, we obtained a list of 338 brain maturation and geometric features from individual T2-weighted MRI data for our subjects. We provided the full list of 510 features and the final list of 338 features in the **Supplemental Materials**. This brain feature vector (1×338) was utilized as input of our proposed OAP-EL model. Meanwhile, an ontology graph can be constructed (**section 2.4**), in which the number of nodes within the graph is equal to the number of features (i.e., 338 features).

## 2.4. Ontology Graph Construction

We designated latent relationship among features through building an ontology graph by utilizing two prior-defined ontologies that respectively define brain parcellation as well as brain maturation and geometry. The brain parcellation ontology describes the whole brain segmented into 9 tissues (e.g., cortical grey matter, white matter, and etc.) [26], and 87 regions [26, 27, 34] (e.g., Frontal lobe, Hippocampus, Corupus, Insula, etc.). The brain maturation and geometry ontology lists six brain maturation metrics, including volumes, and cortical thickness, sulcal depth, curvature, gyrification index, and surface area [26]. To facilitate knowledge sharing, we expressed these two ontologies in the Web Ontology Language (OWL) format, which can be read or visualized by typical OWL processing packages, such as Owlready2 in Python (**Data and Code Availability**).

Utilizing domain knowledge stored in the above-mentioned two ontologies, we constructed an unweighted ontology graph $G$, in which dHCP brain maturation and geometric features (i.e., 338 features from dHCP pipeline) were considered as vertices $V = [v_1, ..., v_n]$, and ontology-derived relationships were edges $E = [e_1, ..., e_n]$. The value of edge $e \in E$ between two vertices $v_i$ and $v_j$ $\forall i, j \in n$ was set to be 1, if two features quantify the same brain maturation and geometric metrics (e.g., the *volume* of left frontal lobe and the *volume* of left occipital lobe) or describe the same brain parcellations (e.g., the volume of *left occipital lobe* and the surface area of *left occipital lobe*), otherwise we set it to 0 (**Figure 2**).

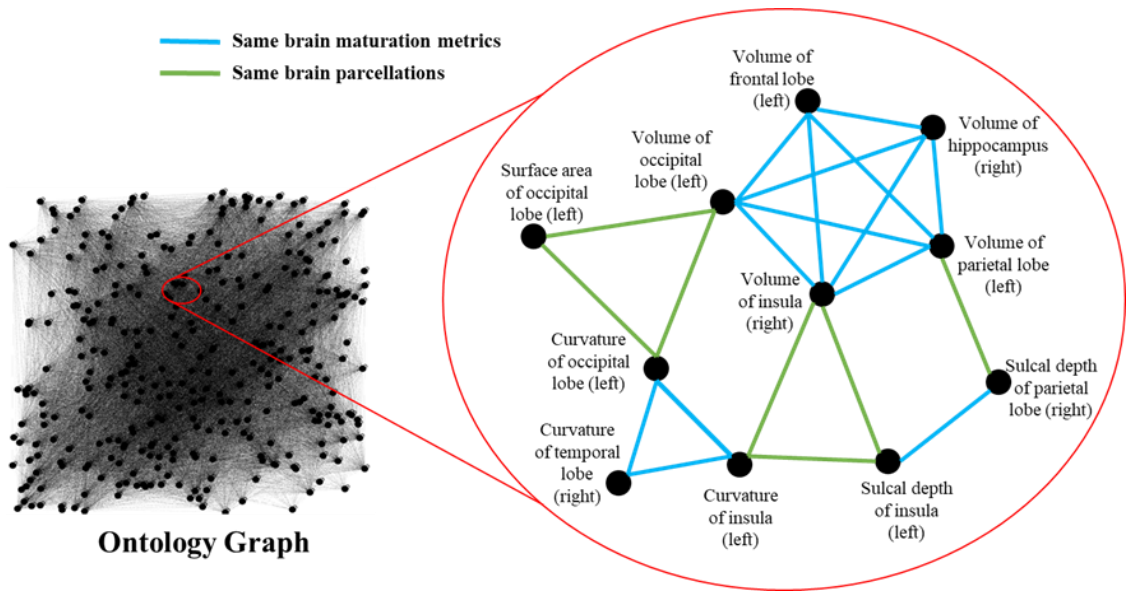

**Figure 2**. Ontology graph with an enlarged subgraph. Based on domain knowledge within two pre-defined ontologies, two vertices are connected if they quantified the same brain maturation and geometric metrics or describe the same brain parcellations, otherwise they are disconnected.

### 2.5. Ontology-guided Attribute Partitioning Ensemble Learning (OAP-EL) model

In contrast to the most commonly used attribute bagging method [14, 15], we conducted ontology graph clustering via a spectral clustering algorithm [36] for feature partitioning. Given our unweighted ontology graph $G = (V, E)$, the graph Laplacian matrix is defined as $L = D - A$, where similarity matrix $A$ of the graph $G$, and $D$ is a degree matrix of graph $G$. Since $L \in R^{n \times n}$ is a positive semidefinite matrix, the eigendecomposition of $L$ is defined as $L = U \Lambda U^{-1}$, where $U \in R^{n \times n}$, whose $i_{th}$ column is the eigenvector $u_i$ of $U$, and $\Lambda \in Z_{\geq 0}^{n \times n}$ is a diagonal matrix whose diagonal elements $\Lambda_{ii} = \lambda_i$ corresponding to its eigenvalue. The spectral clustering algorithm outputs $k$ sets of cluster labels by performing a $k$-means on the first $k$ eigenvectors $U_k$ of $L$, such that, $U_k \in R^{n \times k} \subseteq U$. The number of clusters $k$ is a hyperparameter that can be optimized based on the performance of downstream tasks. We finally partitioned brain maturation and geometric features into $k$ non-overlapping feature subsets.

Using $k$ non-overlapped OAP feature subsets, we built $k$ XGBoost models [37] as base-classifiers. Assume that $c_i \in C\ \forall i \in [1, \ldots, k]$ is the $i_{th}$ OAP feature subset, and $f_i^b \in F^b\ \forall i \in [1, \ldots, k]$ represents $i_{th}$ base-classifier. To train each $f_i^b$, we minimized the loss function:

$$\mathcal{L}(\phi) = \sum_{i=1}^{N} \ell(y_i, \hat{y}_i) + \gamma T + \sum_{j=1}^{T} \frac{1}{2} \eta s_j^2 \quad (2)$$

where $\ell(y_i, \hat{y}_i)$ represents the convex function between true label $y_i$ and prediction $\hat{y}_i$, $\gamma$ and $\eta$ represent the shrinkage parameters for penalizing the model complexity through adjusting the number of leave node j which corresponds to its output of scores $s_j$. Each $f_i^b$ takes the input of a dataset corresponding to a $c_i$ and produces a probabilistic outcome $\hat{p}_i = f_i^b(c_i)\ \forall i \in [1, \ldots, k]$. Hence, $\hat{P} = F^b(C)$ where $\hat{P} = [\hat{p}_1, \ldots, \hat{p}_k]$ represents a set of probabilities from $F^b$ that will be the input for the meta-classifiers.

We used a neural network model as meta-classifier $f^m$ to integrate $k$ probabilities $\hat{p}_i \in \hat{P}\ \forall i \in [1, \ldots, k]$ of $f_i^b \in F^b\ \forall i \in [1, \ldots, k]$. The neural network contains an input layer, followed by a fully connected hidden layer with a Rectified Linear Unit (ReLU) as activation function, and an output layer using a sigmoid function. The final probabilistic outcome $\hat{p}^*$ can be defined as $\hat{p}^* = f^m(\hat{P}) = [1 + exp\left(-\left(0, W\hat{P} + b\right)^+\right)]^{-1}$, where $W$ and $b$ are the weight matrix and bias. To train $f^m$, we minimized the binary cross entropy loss function with $L_2$ norm regularization, which was given by

$$\mathcal{L}^* = -\frac{1}{M} \sum_{i=1}^{M} [y_i \log \hat{p}^* + (1 - y_i) \log(1 - \hat{p}^*)] + \lambda ||W||^2 \quad (3)$$

where $M$ is the sample size, $y_i$ represents the $i_{th}$ class label $\forall i \in [1, ..., M]$, and $\lambda$ is the coefficient of $L_2$ norm regularization, penalizing the weight matrix $W$ to avoid the overfitting problem. For selecting the hyperparameters of $\lambda$ and maximum depth $m$, we tuned the model using a grid search (i.e., $\lambda = [0.001, 0.01, 0.1]$; $m = [2, 4, 6, 8]$). We trained the neural network for the meta-classifiers with 1000 epochs using an Adam optimization algorithm with a learning rate of 0.01, and $\lambda = 0.001$ was chosen for $L_2$ norm regularization.

## 2.6. Internal and External Model Validation

We evaluated the proposed OAP-EL model using both internal and external validation experiments with performance metrics of accuracy, sensitivity, specificity, and area under the receiver operating characteristic (ROC) curve (AUC). To evaluate the model performance on positive minority samples, we plotted Precision-Recall (PR) curves and calculated area under the PR curve (PRAUC). We further reported the mean and standard deviation of performance metrics from 100 experiment replications. For internal validation, we trained and tested the model using CINEPS cohort with a nested Leave-One-Out Cross Validation (LOOCV) strategy, which includes an outer loop and an inner loop. In the outer LOOCV loop, we separated the dataset into training-validation data (*N*-1 samples) and testing data (1 sample) in each of *N* iterations and repeated this process iteratively until all subjects were treated as testing data. Performance metrics were calculated on testing data. In the inner LOOCV loop, the model hyperparameters were optimized using training-validation data (N-1 samples) without seeing any testing data. For external validation, we tested the internally validated model using the unseen independent COEPS cohort.

To handle the data imbalance issue, we applied a data synthesis approach on the training dataset after LOOCV data split. For the CINEPS dataset, we have a smaller number of very preterm infants in the high-risk group compared to those in the low-risk group with the ratio of 1:2. With an imbalanced dataset, machine learning models are prone to become majority class classifiers, i.e., they fail to learn the concepts of the minority class. To overcome this challenge, we first generated new synthetic samples for minority class (for model training only) using synthetic minority over-sampling (SMOTE) method [38]. Specifically, we randomly selected a sample from minority class and obtained its five nearest neighbors. We interpolated new synthetic samples between the selected minor samples and its nearest neighbors. We repeated this process until the ratio of high-risk and low-risk subjects in training dataset was 1:1. Next, we implemented edited nearest neighbor [39] method (ENN). We also selected a random sample in the majority class and obtained its five nearest neighbors. If the class of selected sample was different from its five nearest neighbors, these samples were removed regardless their class labels in majority or minority class. This process was repeated until the ratio of high-risk and low-risk subjects in training dataset was 1:1. The procedure was referred to as SMOTE-ENN method [39] with additional details illustrated in the **Supplementary Table 1**.

We compared our proposed model with 1) traditional single-channel machine learning models, including *K*-Nearest Neighbor (KNN) [40], Logistic Regression (LR) [41], Support Vector Machine (SVM) [42], Decision Tree (DT) [43], Random Forest (RF) [44], Neural Network (NN) [45]; 2) peer ensemble learning models, including Voting [46], Bagging [47], Stacking [48], and Attribute Bagging-Ensemble Learning (AB-EL); and 3) multi-channel neural networks (mNN) that was developed in our prior study [49]. The detailed implementation of all the models can be found in **Supplemental Materials**. All the machine learning experiments were performed in a workstation with a processor with Intel(R) Core(TM) i5-10600KF CPU at 4.10GHz, 8 GB RAM, and a NVIDIA GeForce GTX 1660 SUPER GPU. Experiment coding was conducted using Python 3.7, TensorFlow 2.3.0, and Scikit-Learn 0.24.1.

## 2.7. Identification of Discriminative Features

We identified and reported the top discriminative brain geometric features that contributed most to the prediction of cognitive deficit by utilizing a two-level feature importance ranking strategy. Within our OAP-EL model, suppose that $W$ represents the connection weights of the meta-classifier, and weight $w_i \in W \ \forall i \in [1, ..., k]$ corresponds to the $i_{th}$ base-classifier. Let $\beta_{ij} \in B_i \forall j \in [1, ..., |c_i|], c_i \in C$ be the $j_{th}$ feature importance score of $i_{th}$ XGBoost base-classifier using the information gain [50], where $|c_i|$ is the size of features within the $i_{th}$ base-classifier. The global ranking score of a brain maturation and geometric feature is defined as $\frac{w_i}{\sum_{i=1}^{k} w_i} \cdot \frac{\beta_{ij}}{max(B_i)}$.

## 2.8. Statistical Analysis

To examine demographic differences between the groups of high-risk and low-risk infants, we used two-sample Student's t-test to compare means for continuous variables, includes birth weight (BW), gestational age at birth (GA), postmenstrual age (PMA) at scan and cognitive score, and Pearson's chi-squared test to compare gender distribution. To compare the different prognostic models, we also used the nonparametric Wilcoxon test. A p-value less than 0.05 was considered statistically significant for all inference testing. All statistical analyses were conducted in R-4.0.3 (RStudio, Boston, MA, USA).

## 3. Results

### 3.1. Finding the Optimal Number of Feature Clusters $k$

The number of feature clusters was optimized using the internal validation cohort. Specifically, we tested the numbers of clusters $k$ with empirical values from 1 to 100 in increments of 1. For each $k$, we repeated nested LOOCV 100 times to evaluate prediction performance. **Figure 3(A)** shows the mean AUC with various numbers of clusters $k$. According to the highest mean AUC, we set the optimal numbers of feature clusters to be 6 in the following experiments. Further, we applied silhouette score to evaluate the goodness of ontology-guided clustering. Silhouette score is a measure of how similar an object (i.e., feature) is to its own cluster compared to other clusters. It ranges from −1 to +1, where a higher value indicates that objects are clustered better. We calculated silhouette score for each feature in our ontology graph, and then averaged silhouette scores across all features. A detailed calculation of the mean silhouette score is included in the Supplemental Material. **Figure 3(B)** shows that the optimal number of feature cluster (k=6) appears to have the highest silhouette score compared to other number of feature cluster, demonstrating that our proposed OAP is able to partition features into better clusters by only using prior domain knowledge. Besides, the Ontology graph (k=6) using spectral clustering is shown in **Figure 4**. The bubbles and link represent the sample nodes (i.e., features) and edges to illustrate some clustered feature examples.

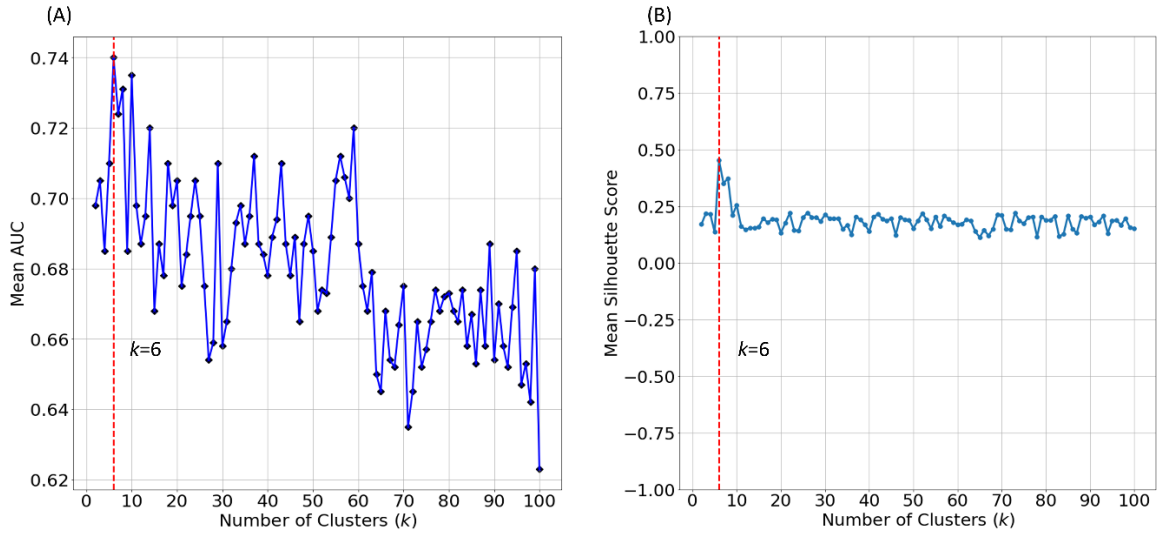

**Figure 3**. Optimization of the number of feature clusters k for early prediction of cognitive deficits in very preterm infants. (A) mean AUC of the predictive model with varying numbers of clusters. (B) mean silhouette scores of varying numbers of clusters.

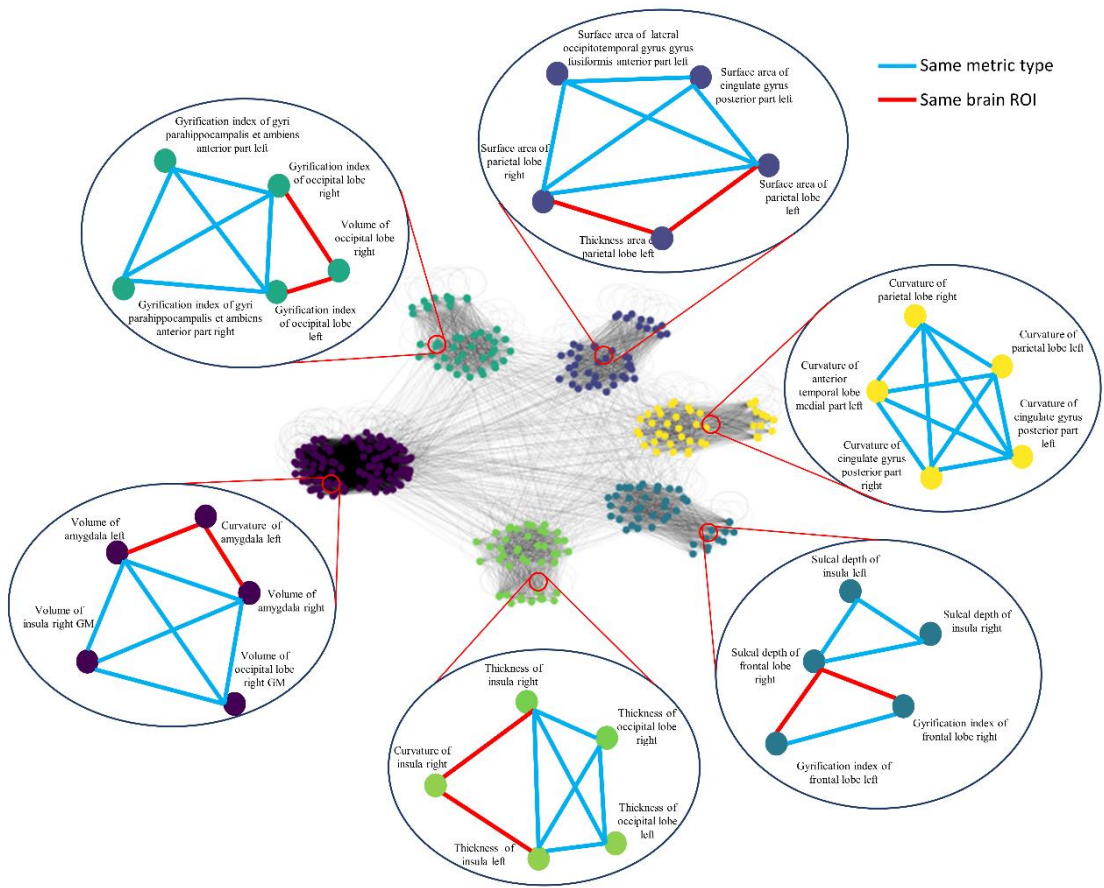

**Figure 4**. Ontology graph partitioned using spectral clustering algorithm into the optimal number of clusters (k = 6). Sample nodes and edges in each cluster are displayed in the bubble. Nodes are colored according to their clusters. Edge colors are added in indicate same metric type (blue) or same brain ROI (red).

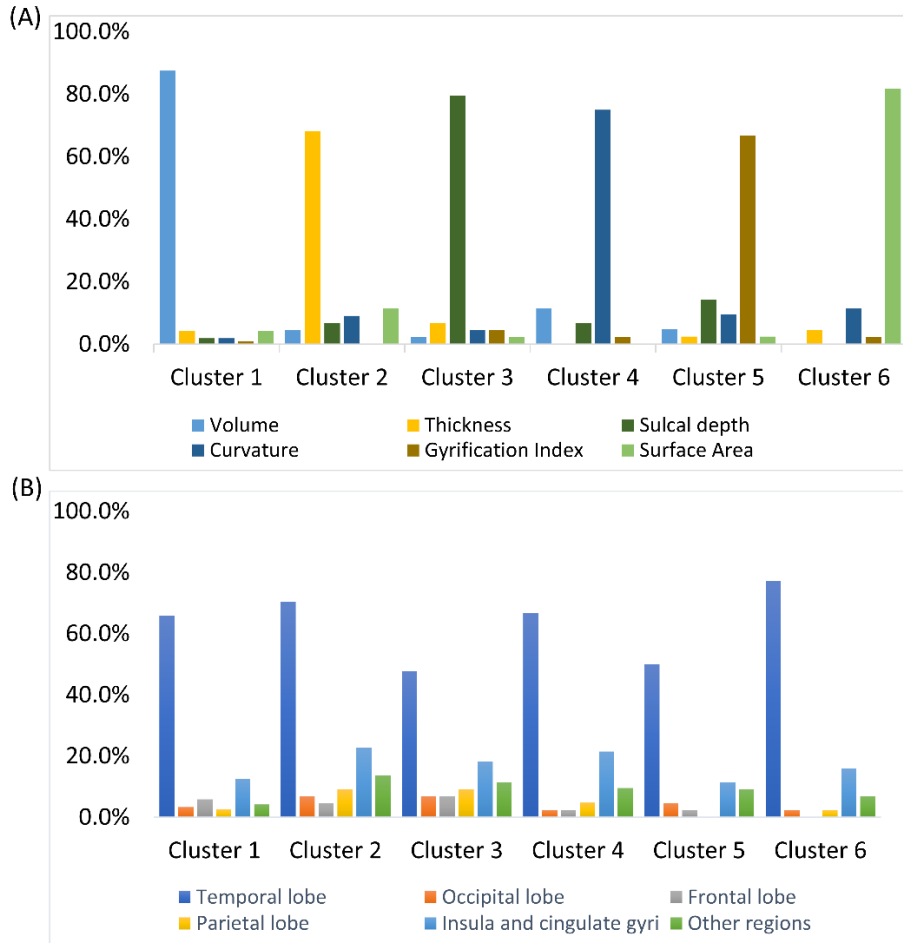

**Figure 5.** The distribution of (A) brain metrics and (B) brain regions among feature clusters. For those brain regions (hippocampus, amygdala, cerebellum, cerebrospinal fluid) that contain very few features, we merged them together as 'other regions' for visualization purpose.

We further analyzed the feature clusters derived from our OAP approach. We illustrated the distribution of the feature types and representative ROIs in each one of six feature clusters. (**Figure 5**) We noted from **Figure 5(A)** that each feature cluster (generated from ontology graph) is dominated by one feature type, and none of feature cluster contains a single feature type. This clearly demonstrates that different types of brain metrics have complementary information, which may improve the discriminative power of base-classifiers. On the other hand, from **Figure 5(B)**, we observed that all feature clusters were dominated by features related to temporal lobe. This is simply because that the adopted brain parcellation ontology divided the temporal lobe into more sub-regions than other major brain lobes and regions, resulting more features from temporal lobe.

### 3.2. Internal Validation with CINEPS Cohort

We included 207 very preterm infants (mean (SD) GA of 29.4 (2.4) weeks; birth weight of 1288.3 (434.5) grams) who completed 2-year cognitive assessments from CINEPS cohort for internal validation. (**Table 1**) We defined infants with a Bayley III cognitive score less than or equal to 85 (i.e., 1 SD below the mean score) as the high-risk group (N=69) and those with a cognitive score greater than 85 as the low-risk group (N=138) for developing cognitive deficits. The sample ratio between the high- and low-risk groups was 1:2 in the CINEPS cohort. The high-risk group had a mean GA of 28.7 (2.7) weeks, PMA of 42.5 (1.3) weeks, birth weight of 1184.1 (489.3) grams, and 43 of 69 (62.3%) subjects were male. The low-risk group had a mean GA of 29.7 (2.2) weeks, PMA of 42.4 (1.3) weeks, birth weight of 1340.5 (397.3) grams, and 67 of 138 (48.6%) subjects were male. Between groups, there was significant difference in birth weight (p=0.02), gestational age at birth (p<0.001) and cognitive scores (p<0.001); and no significant difference in sex (p=0.08) and postmenstrual age at scan (p=0.48).

**Table 1**. Demographics summary of CINEPS cohort.

|  | CINES Cohort | | |
| --- | --- | --- | --- |
|  | Low-risk (N=138) | High-risk (N=69) | p-value |
| **Male sex; Number (%)** | 67 (48.6%) | 43 (62.3%) | 0.08 |
| **PMA at scan in weeks; Mean (SD)** | 42.4 (1.3) | 42.5 (1.3) | 0.48 |
| **GA in weeks; Mean (SD)** | 29.7 (2.2) | 28.7 (2.7) | **<0.001** |
| **Birth weight in grams; Mean (SD)** | 1340.5 (397.3) | 1184.1 (487.3) | **0.02** |
| **Cognitive assessment at 2 years corrected age; Mean (SD)** | 99.2 (8.8) | 76.0 (10.7) | **<0.001** |

### *3.2.1. Effects of Individual Ontologies on OAP-EL*

We first investigated the effects of individual ontologies on our early prediction task. We constructed three ontology graphs using (1) brain metric ontology, (2) brain parcellation ontology, and (3) combined ontologies, respectively. Feature clustering was performed on all ontology graphs, and OAP-EL models were developed using internal cohort. As shown in **Table 2**, the OAP-EL model using combined ontologies achieved a mean AUC of 0.74, significantly higher than the OAP-EL models using either brain metric ontology (AUC=0.69, $p<0.001$) or brain parcellation ontology (AUC=0.61, $p<0.001$). This demonstrates that individual brain ontologies have their own discriminative power, and combining ontologies together is able to further increase the overall discriminative power by using complementary knowledge from individual ontologies.

**Table 2**. Model performance of the proposed OAP-EL using brain metric ontology, brain parcellation ontology, and combined ontologies.

| Ontologies | Accuracy (%) | Sensitivity (%) | Specificity (%) | AUC |
|---|---|---|---|---|
| Brain metric ontology | 69.1 (3.1) | 67.3 (3.8) | 70.3 (2.9) | 0.69 (0.03) |
| Brain parcellation ontology | 61.7 (3.4) | 61.9 (4.2) | 63.5 (3.2) | 0.61 (0.04) |
| **Combined ontologies** | **71.3 (1.9)** | **70.6 (2.0)** | **72.6 (1.7)** | **0.74 (0.03)** |

*3.2.2. OAP-EL Outperforms Traditional Machine Learning Models*

**Figure 6** shows the performance comparison among our proposed OAP-EL model and traditional machine learning models in the detection of very preterm infants at high-risk for moderate/severe cognitive deficits. When compared to the best performing traditional machine learning model, SVM, our proposed OAP-EL model demonstrated a significantly higher accuracy by 8.2% ($p<0.001$), sensitivity by 9.0% ($p<0.001$), specificity by 7.5% ($p<0.001$), and AUC by 0.1 ($p<0.001$).

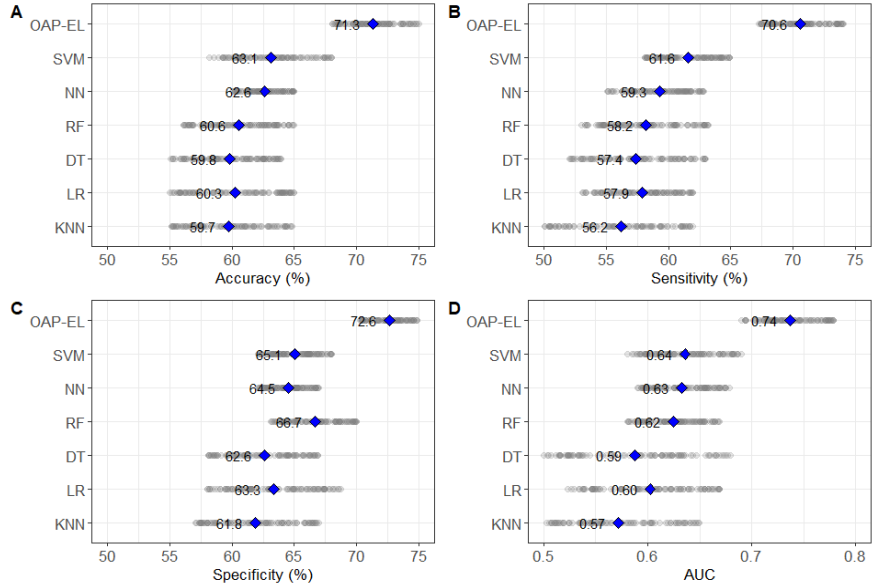

**Figure 6.** Internal validation of the proposed OAP-EL and traditional machine learning models on (A) Accuracy, (B) Sensitivity, (C) Specificity, and (D) AUC. KNN: *K*-nearest neighbor; LR: Logistic Regression; SVM: Support Vector Machine; DT: Decision Tree; RF: Random Forest; NN: Neural Network; OAP-EL: Ontology-guided Attribute Partitioning Ensemble Learning. The highlight points indicate the mean value of measures.

Both ROC and PR curves demonstrate that our OAP-EL model had a superior prediction performance. (**Figure 7**) Our proposed OAP-EL achieved the better AUC and PRAUC than other traditional machine learning models.

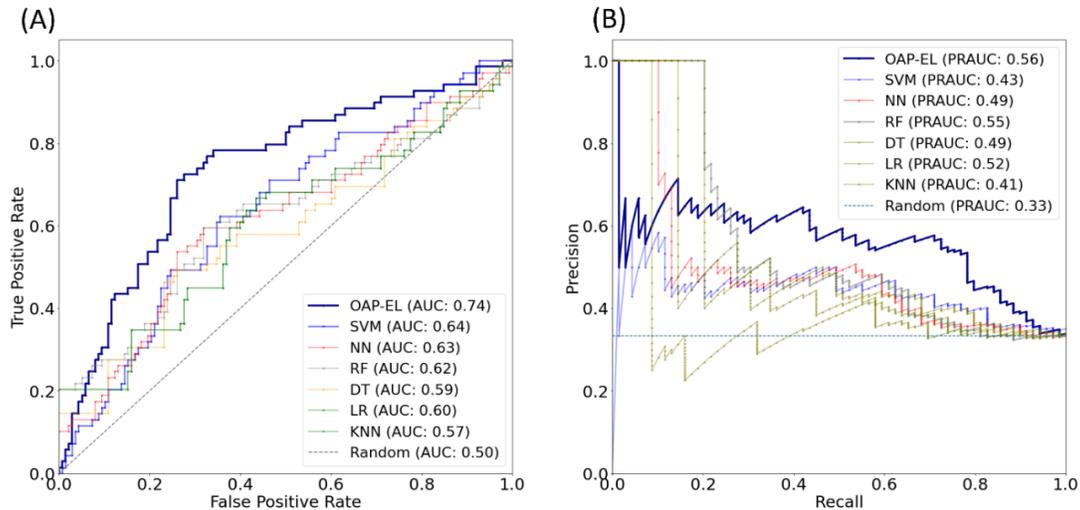

**Figure 7.** Model performance comparison between the OAP-EL method and traditional machine learning methods for internal validation using (A) receiver operating characteristic curves, and (B)

precision-recall curves. AUC-Area under the ROC curve; PRAUC-Area under the Precision-Recall curve.

### 3.2.3 OAP-EL Outperforms Peer Ensemble Learning Models

We compared the proposed OAP-EL model with several peer ensemble learning models, including Voting, Bagging, and Stacking without/with attribute bagging method. The prediction performance is shown in **Table 3**. Our proposed OAP-EL model achieved best prediction performance among peer ensemble learning models. The proposed model significantly improved the prediction performance over the AB-EL model by 3.8% in accuracy (p<0.001), 5.4% in sensitivity (p<0.001), 4.2% in specificity, and 0.05 in AUC (p<0.001). **Figure 8** illustrates both ROC and PR curves of ensemble learning models. While ROC curves shows that our OAP-EL model had a better prediction performance, PR curves demonstrate the superior performance of our OAP-EL model over peer ensemble learning models with a larger margin.

**Table 3.** Internal validation of the proposed OAP-EL and ensemble learning models on Accuracy, Sensitivity, Specificity, and AUC. AB- EL: Attribute Bagging Ensemble Learning, OAP-EL: Ontology-guided Attribute Partitioning Ensemble Learning. Experiment results are represented as mean (SD).

| Model | Accuracy (%) | Sensitivity (%) | Specificity (%) | AUC |
|---|---|---|---|---|
| Voting | 62.3 (2.5) | 57.5 (2.8) | 64.7 (2.3) | 0.62 (0.02) |
| Bagging | 63.4 (3.1) | 62.5 (3.1) | 67.1 (2.6) | 0.63 (0.03) |
| Stacking | 63.8 (4.2) | 61.6 (2.1) | 67.3 (2.7) | 0.63 (0.03) |
| AB-EL | 67.5 (3.8) | 65.2 (4.4) | 68.4 (3.5) | 0.69 (0.03) |
| **OAP-EL** | **71.3 (1.9)** | **70.6 (2.0)** | **72.6 (1.7)** | **0.74 (0.03)** |

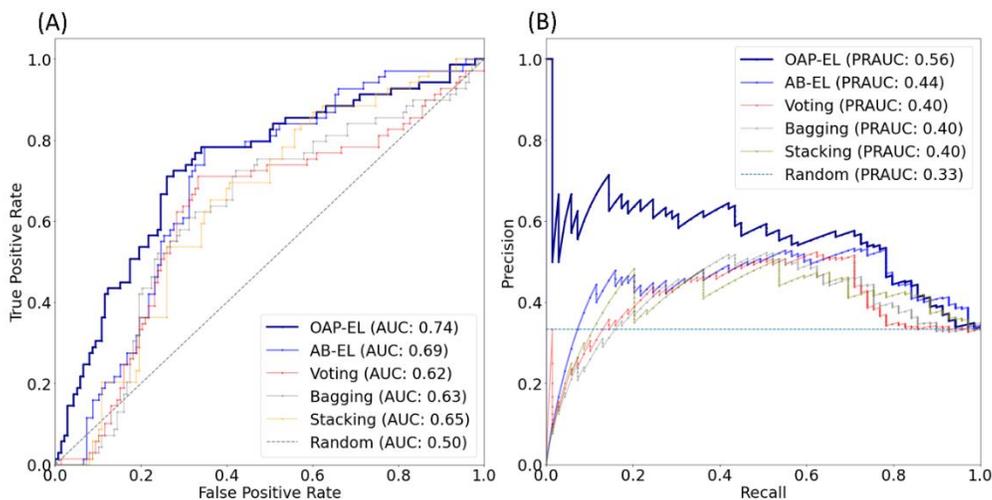

**Figure 8.** Model performance comparison between the OAP-EL method and peer ensemble learning methods for internal validation using (A) receiver operating characteristic curves, and (B) Precision-Recall curves. AUC-Area under the ROC curve; PRAUC-Area under the Precision-Recall curve.

To evaluate the effect of spectral clustering algorithm on the ontology graph, we further compared the model performance using OAP-derived feature clusters ($k=6$) and random feature clusters ($k=6$) (**Figure 9**). The model using OAP-derived feature clusters ($k=6$) achieved 71.3% accuracy, 70.6% sensitivity, 72.6% specificity, and 0.74 AUC. It was significantly better than the model using random feature clusters ($k=6$) that achieved an accuracy of 64.1% (p<0.001), sensitivity of 63.6% (p<0.001), specificity of 67.2% (p<0.001), and AUC of 0.66 (p<0.001).

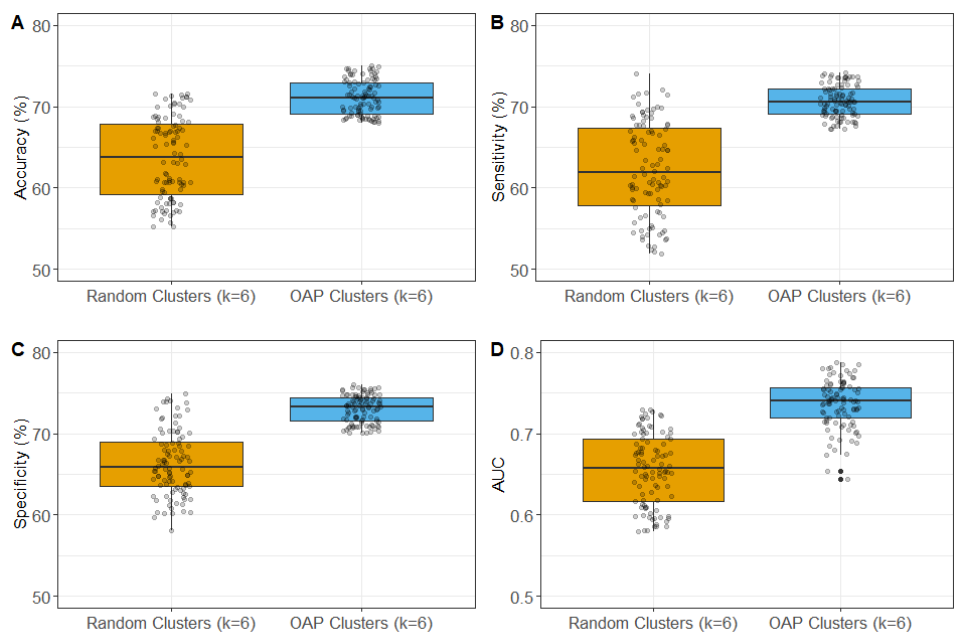

**Figure 9**. Prediction performance comparison of models using OAP-derived feature clusters ($k=6$) and random feature clusters ($k=6$).

### 3.2.4. OAP-EL Outperforms OAP-enhanced Multi-Channel Neural Network (OAP-mNN)

We compared the proposed OAP-EL model with OAP-mNN model (**Supplemental Materials**). We observed that the proposed OAP-EL achieved significantly better prediction performance than the OAP-mNN by 5.4% (p<0.001) in accuracy, 4.8% (p<0.001) in sensitivity, 2.2% (p<0.001) in specificity, and 0.04 (p<0.001) in AUC (**Figure 10**). The PR curves in **Figure 11** demonstrates that our OAP-EL model had a significantly better prediction performance than the OAP-mNN model, even though ROC curves of both OAP-EL and OAP-mNN appear comparable.

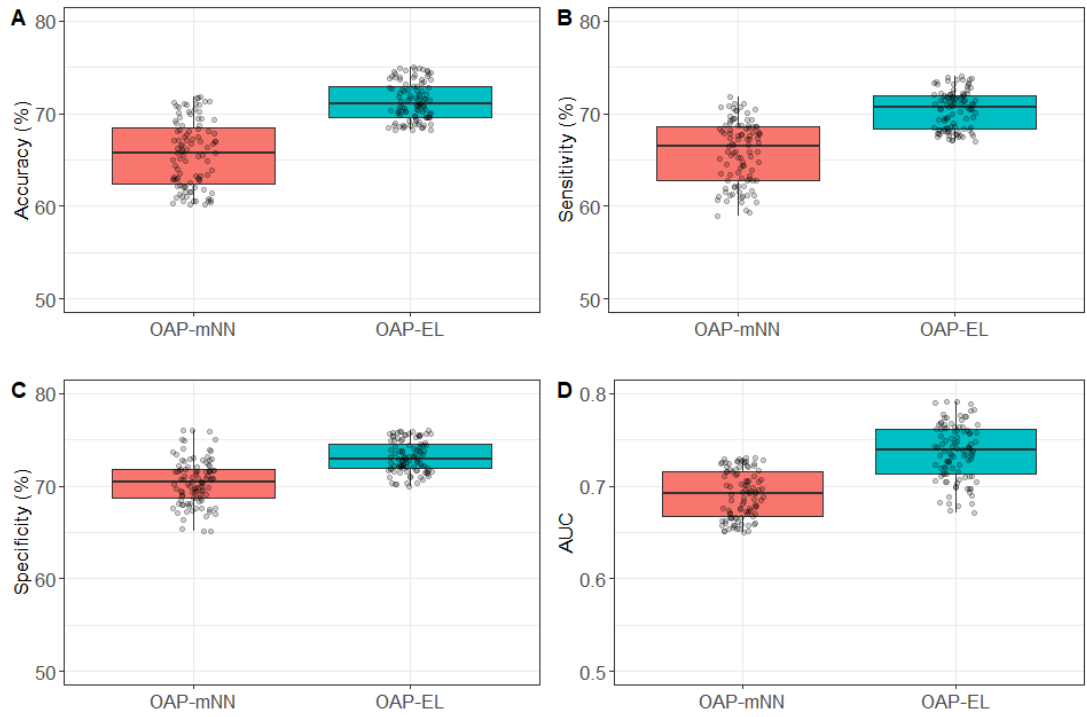

**Figure 10.** Internal risk prediction performance of cognitive deficits comparison with Multi-Channel Neural Network models on (A) Accuracy, (B) Sensitivity, (C) Specificity, and (D) AUC. OAP-mNN: Ontology-guided Attribute Partitioning Multi-Channel Neural Network; OAP-EL: Ontology-guided Attribute Partitioning Ensemble Learning.

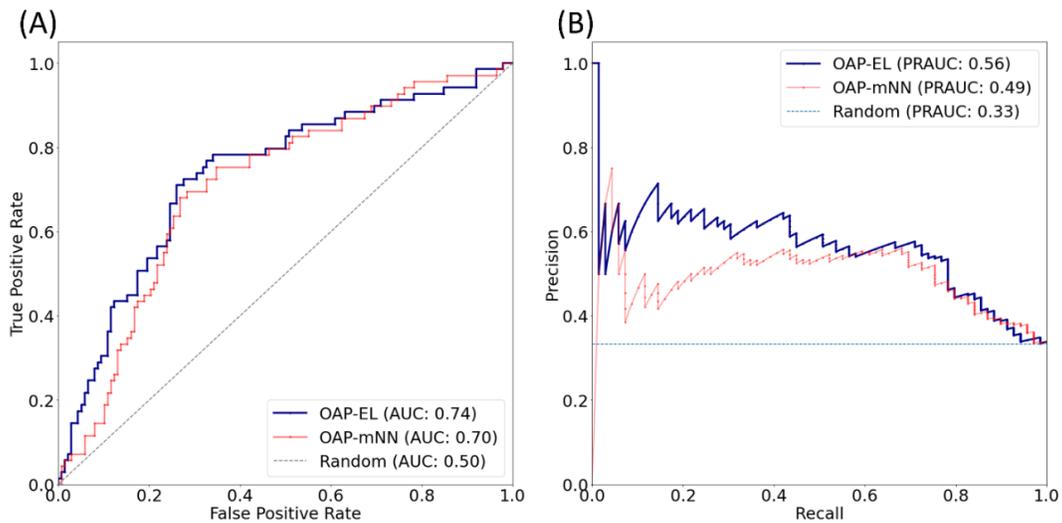

**Figure 11.** Model performance comparison between the OAP-EL method and multi-channel neural networks for internal validation using (A) receiver operating characteristic (ROC) curves, and (B) Precision-Recall curves. AUC-Area under the ROC curve; PRAUC-Area under the Precision-Recall curve.

### 3.2.5. An Ablation Study on the Impact of Individual Brain Metric Types

In this section, we conducted an ablation study to assess the impact of individual brain maturation and geometry metric types. Specifically, we removed one type of metrics and only utilized the rest of five types to develop and evaluate the proposed OAP-EL model. For example, we excluded all brain volume metric related features and retained all other features in one trial. We performed a total of six trials, in each of which we excluded only one type of metrics. **Table 4** shows the prediction performance of OAP-EL model across all trials. The OAP-EL model achieved the best accuracy in Trial 2, where we removed all features related to gyrification index. This indicates that the impact of gyrification index is the least among all six metric types. On the other hand, in Trial 4, where all curvature related features were excluded, the OAP-EL model had the lowest mean prediction accuracy of 66.5%, compared to a mean accuracy of 71.3% when all features were included. This means that curvature related features might have the largest impact on model accuracy. Overall, we observed that all six metric types have their own discriminative power on early prediction of cognitive deficits in very preterm infants.

**Table 4.** An ablation study to assess the impact of individual metric types (m1-m6) on prediction performance for cognitive deficits. In each trial, one metric type was excluded and the rest of five types were utilized to develop and evaluate the proposed OAP-EL model.

| Trial # No. | $m_1$ | $m_2$ | $m_3$ | $m_4$ | $m_5$ | $m_6$ | Accuracy (%) | Sensitivity (%) | Specificity (%) | AUC |
|---|---|---|---|---|---|---|---|---|---|---|
| 1 | ✗ | ✓ | ✓ | ✓ | ✓ | ✓ | 66.7 (2.9) | 65.1 (3.6) | 68.4 (3.0) | 0.68 (0.02) |
| 2 | ✓ | ✗ | ✓ | ✓ | ✓ | ✓ | 69.4 (3.4) | 68.5 (3.5) | 70.5 (3.1) | 0.72 (0.03) |
| 3 | ✓ | ✓ | ✗ | ✓ | ✓ | ✓ | 67.2 (4.2) | 64.7 (4.1) | 67.2 (2.4) | 0.65 (0.03) |
| 4 | ✓ | ✓ | ✓ | ✗ | ✓ | ✓ | 66.5 (3.6) | 65.9 (4.5) | 66.4 (3.3) | 0.67 (0.04) |
| 5 | ✓ | ✓ | ✓ | ✓ | ✗ | ✓ | 68.5 (3.1) | 67.4 (3.8) | 69.3 (2.8) | 0.70 (0.02) |
| 6 | ✓ | ✓ | ✓ | ✓ | ✓ | ✗ | 66.8 (3.9) | 64.0 (4.7) | 68.1 (3.9) | 0.65 (0.03) |
| **All** | ✓ | ✓ | ✓ | ✓ | ✓ | ✓ | **71.3 (1.9)** | **70.6 (2.0)** | **72.6 (1.7)** | **0.74 (0.03)** |

$m_1$: volume; $m_2$: gyrification index; $m_3$: thickness; $m_4$: curvature; $m_5$: surface area; $m_6$: sulcal depth

### 3.3. External Validation with COEPS Cohort

We included 69 very preterm infants with mean (SD) GA of 28.2 (2.4) and birth weight of 1123.6 (400.1) from the independent COEPS cohort for external validation. (**Table 5**) Like the CINEPS cohort, we used a Bayley III 85 as cutoff score to stratify the risk of cognitive deficits, in which infants with a score less than or equal to 85 were defined as a high-risk group (N=10), while infants with a score greater than 85 were defined as a low-risk group (N=59). The sample ratio between high- and low-risk group was ~1:6 for the COEPS cohort. The high-risk group of infants had a mean GA of 26.8 (2.7) weeks, PMA at MRI scan of 40.4 (0.7) weeks, birth weight of 904.5 (358.8) grams, and 5 of 10 (50.0%) subjects were male. The low-risk group of infants had a mean (SD) GA of 28.5 (2.2) weeks, PMA at MRI scan of 40.3 (0.5) weeks, birth weight of 1160.8 (397.6) grams, and 25 of 59 (42.3%) subjects were male. As with the internal cohort, we observed significant differences between the high-risk vs. low-risk group in birth weight (p=0.06), GA (p=0.04) and cognitive scores (p<0.001) but no significant differences in sex (p=0.20) or PMA at MRI scan (p=0.78).

**Table 5**. Demographics summary of COEPS cohort.

|  | COEPS Cohort | | |
| --- | --- | --- | --- |
|  | Low-risk (N=59) | High-risk (N=10) | p-value |
| **Male sex; Number (%)** | 34 (57.6%) | 5 (50.0%) | 0.20 |
| **PMA at scan in weeks; Mean (SD)** | 40.3 (0.5) | 40.4 (0.7) | 0.78 |
| **GA in weeks; Mean (SD)** | 28.5 (2.2) | 26.8 (2.7) | **0.04** |
| **Birth weight in grams; Mean (SD)** | 1160.8 (397.6) | 904.5 (358.8) | **<0.001** |
| **Cognitive assessment at 2 years corrected age; Mean (SD)** | 102.5 (10.7) | 71.0 (12.4) | **<0.001** |

The final trained models (using the internal cohort) were tested using this external cohort and their performance is shown in **Table 6**. We noticed that the comparison results in external validation exhibited a similar trend to the results in the internal validation. (**Figure 12**) The external validation further illustrate that the proposed OAP-EL model was able to outperform other traditional machine learning and peer ensemble learning models in unseen data from another study site.

**Table 6.** External validation comparison of the proposed OAP-EL and ensemble learning models to assess Accuracy, Sensitivity, Specificity, and AUC. KNN: *K*-nearest neighbor; LR: Logistic Regression; SVM: Support Vector Machine; DT: Decision Tree; RF: Random Forest; NN: Neural Network; OAP-mNN: Ontology-guided Attribute Partitioning Multi-Channel Neural Network; AB-EL: Attribute Bagging Ensemble Learning; OAP-EL: Ontology-guided Attribute Partitioning Ensemble Learning.

| Model | Accuracy (%) | Sensitivity (%) | Specificity (%) | AUC |
|---|---|---|---|---|
| KNN | 44.9 | 30.0 | 47.5 | 0.52 |
| LR | 56.5 | 40.0 | 59.3 | 0.56 |
| DT | 52.2 | 30.0 | 55.9 | 0.55 |
| RF | 59.4 | 50.0 | 61.0 | 0.60 |
| NN | 63.8 | 50.0 | 66.1 | 0.62 |
| SVM | 66.7 | 60.0 | 67.8 | 0.63 |
| Voting | 60.9 | 50.0 | 62.7 | 0.61 |
| Bagging | 56.5 | 40.0 | 59.3 | 0.62 |
| Stacking | 62.3 | 50.0 | 64.4 | 0.62 |
| OAP-mNN | 66.7 | 60.0 | 67.8 | 0.66 |
| AB-EL | 68.1 | 60.0 | 69.4 | 0.68 |
| **OAP-EL** | **71.0** | **70.0** | **71.2** | **0.71** |

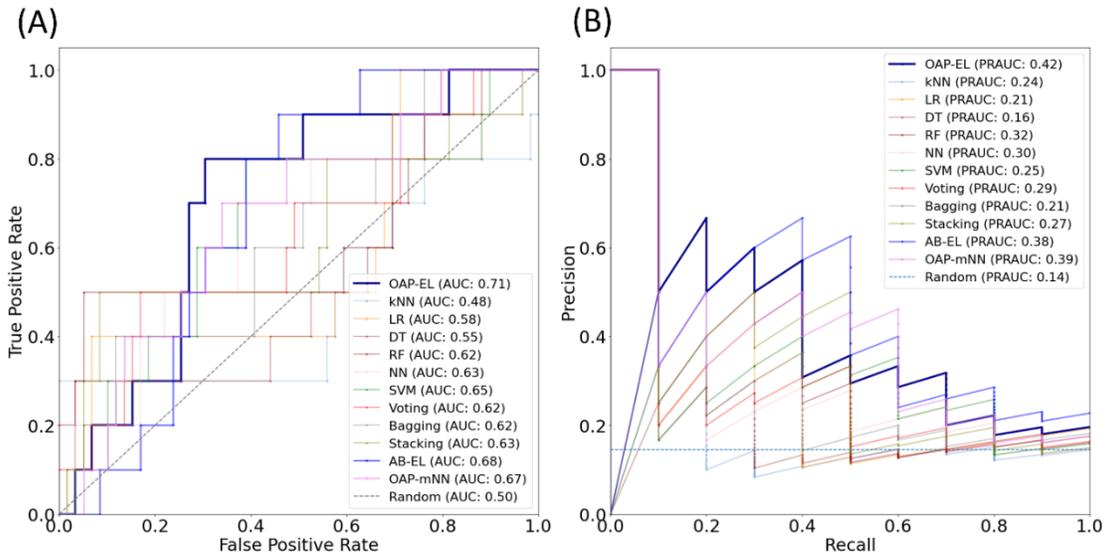

**Figure 12.** Model performance comparison between the OAP-EL method and all alternative methods for external validation using (A) receiver operating characteristic (ROC) curves, and (B) Precision-Recall curves. AUC-Area under the ROC curve; PRAUC-Area under the Precision-Recall curve.

### 3.4. Most Discriminative Brain Maturation and Geometric Features

To identify which features contributed most of the variance for predicting cognitive deficits, we ranked all brain maturation and geometric features using the two-level feature ranking method (**Materials and Methods**). **Table 7** displays the top 15 predictive brain maturation and geometric features identified by our OAP-EL model as well as their ranking scores. The thickness of the insula region within the right hemisphere was ranked as the most predictive feature. This was followed by sulcal depth measure for the anterior part of medial and inferior temporal gyri from the left hemisphere. In terms of metric types, we noted that thickness and sulcal depth were two frequent types (9 out of 15) among these top features, even though the other metric types were also represented. We further visualized the top brain regions in **Figure 13**.

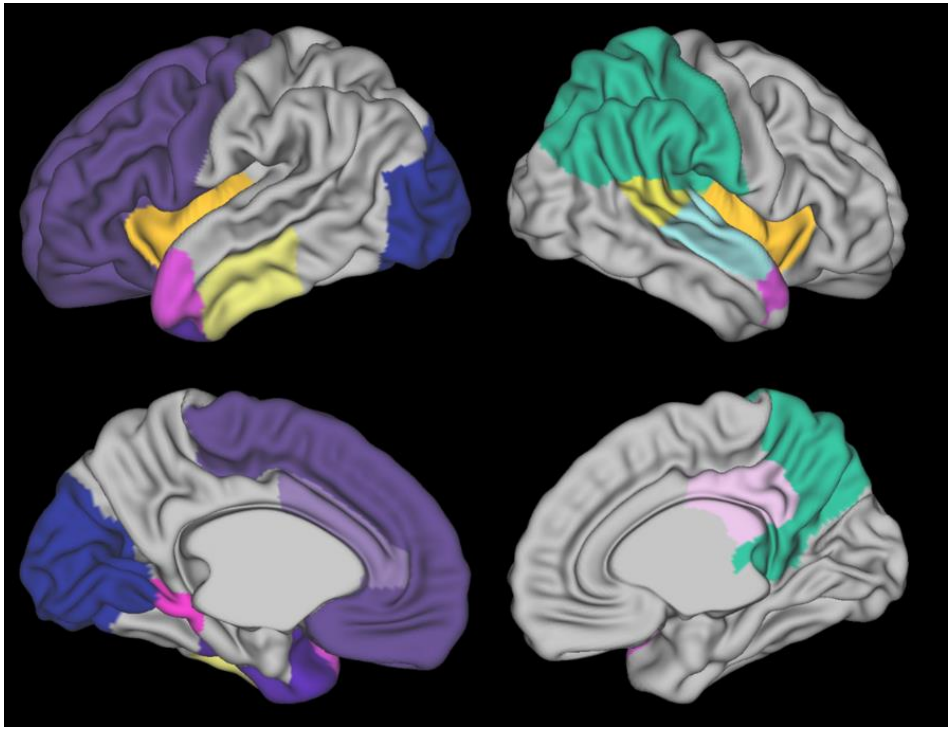

**Figure 13.** Visualization of top 15 most discriminative brain regions learned by the Ontology-guided Attribute Partitioning Ensemble Learning (OAP-EL) model.

**Table 7.** Most Prominent Brain Maturation and Geometric Features Ranking.

| Metric Types | Brain Regions | Ranking Score |
|---|---|---|
| Thickness | Insula right | 0.76 |
| Sulcal depth | Medial and inferior temporal gyri anterior part left | 0.74 |
| Sulcal depth | Cingulate gyrus anterior part left | 0.72 |
| Thickness | Superior temporal gyrus posterior part right | 0.70 |
| Sulcal depth | Gyri parahippocampalis et ambiens posterior part left | 0.68 |
| Curvature | Superior temporal gyrus middle part right | 0.66 |
| Curvature | Temporal lobe left (merged region) | 0.64 |
| Surface area | Anterior temporal lobe medial part left | 0.61 |
| Sulcal depth | Cingulate gyrus posterior part right | 0.60 |
| Volume | Parietal lobe right GM | 0.58 |
| Sulcal depth | Lateral occipitotemporal gyrus fusiformis anterior part left | 0.56 |
| Gyrification index | Anterior temporal lobe lateral part right | 0.55 |
| Thickness | Insula left | 0.52 |
| Gyrification index | Occipital lobe left | 0.49 |
| Thickness | Frontal lobe left | 0.47 |

## 4. Discussion

In this paper, we proposed a novel OAP approach for feature partitioning and developed an OAP-EL model for early prediction of cognitive deficits at 2 years corrected age in very preterm infants using brain maturation and geometric features obtained at term-equivalent age. The model was comprehensively evaluated using internal and external validations with two independent very preterm infant cohorts. Our proposed OAP-EL achieved an accuracy of 71.3%, sensitivity of 70.6%, specificity of 72.6%, and AUC of 0.74 in internal validation; and an accuracy of 71.0%, a sensitivity of 70.0%, a specificity of 71.2%, and AUC of 0.71 in external validation.

### 4.1. Ensemble of Classifiers vs. Individual Classifiers

Early prediction of cognitive deficits for very preterm infants continues to be a particularly challenging task in the clinical setting. Prognostic classifiers can be trained with different feature sets, and each prediction classifier has its own strengths and weaknesses. Therefore, it is natural to expect that a learning method that takes advantages of multiple classifiers would lead to superior performance. To this end, ensemble learning aims to integrate multiple classifiers to complement each other's weaknesses, therefore rendering better performance over each individual classifier

[12]. The intuitive explanation of why ensemble learning works is that our human nature seeks the wisdom of crowds in making a complex decision. An example of such a decision is matching a medical treatment to a particular disease [51-55]. Theoretically, several reasons explain why ensemble learning works, including avoiding overfitting, greater computational efficiency, and hypothesis reinforcement[56, 57]. Our results in both internal and external validation experiments demonstrated that the ensemble of classifiers achieved significantly better prediction performance than individual classifiers.

**4.2. Ensemble Learning with Ontology-guide Feature Partitioning vs. with Random Feature Bagging**

The diversity of both features and classifiers plays a key role and it is a necessary and sufficient condition in building a powerful ensemble model. A diverse set of classifiers in the base-classifier library can be trained using a diverse set of features. Most widely used feature subset partitioning schemes (e.g., random feature bagging) [14, 15] randomly draw feature subset from the entire feature set, which neglects prior domain knowledge and latent relationship among features. In this study, for the first time, we proposed to integrate prior domain knowledge, expressed in ontologies, into feature a partitioning scheme. We demonstrated that the proposed ontology guided attribute partitioning-based ensemble model produced significantly better prediction performance than the classic attribute bagging-based classifier in both internal and external validations.

*Empirical Insights of Classifier Ensembles via Kappa-Error Diagram*

We empirically explain the reasoning why our proposed OAP-ensemble model is better by depicting a kappa-error diagram, a visualization tool for classifier ensembles [58]. The kappa-error diagram is a scatterplot of all pairs of classifiers in the base-classifier library of an ensemble model (i.e., each pair of base-classifiers is represented as a point on the graph). The *x*-coordinate of the point is a measure of *diversity* (denoted as K) between the outputs of the two classifiers. The pairwise K is defined as $K = \frac{2(ab-bc)}{(a+b)(c+d)+(a+c)(b+d)}$, where *a* is the proportion of instances correctly classified by both base-classifiers, *b* is the proportion correctly classified by the first base-classifier but misclassified by the second one, *c* is the proportion misclassified by the first base-classifiers but correctly classified by the second one, and $d$ is the proportion misclassified by both base-classifiers. The lower the K value, the more different the classifiers, and the higher chance to fill in each other's weakness, therefore resulting in better classifier ensembles. The *y*-coordinate of the point is the averaged misclassification rate of the pair of classifiers. The ensemble model with $k$ base-classifiers is represented by a "cloud" of $\frac{k(k-1)}{2}$ points, and $\frac{100k(k-1)}{2}$ points with 100 times repetition of LOOCV. Better ensembles will be the ones with a "cloud" of points near the left bottom corner of the graph (i.e., high diversity and low individual error). **Figure 14** shows that our proposed OAP-EL model produces more accurate and more diverse individual classifiers, since the "clouds" of OAP-EL are more to the left bottom corner of the graph. This indicates the key to the better overall performance we see with OAP-EL.

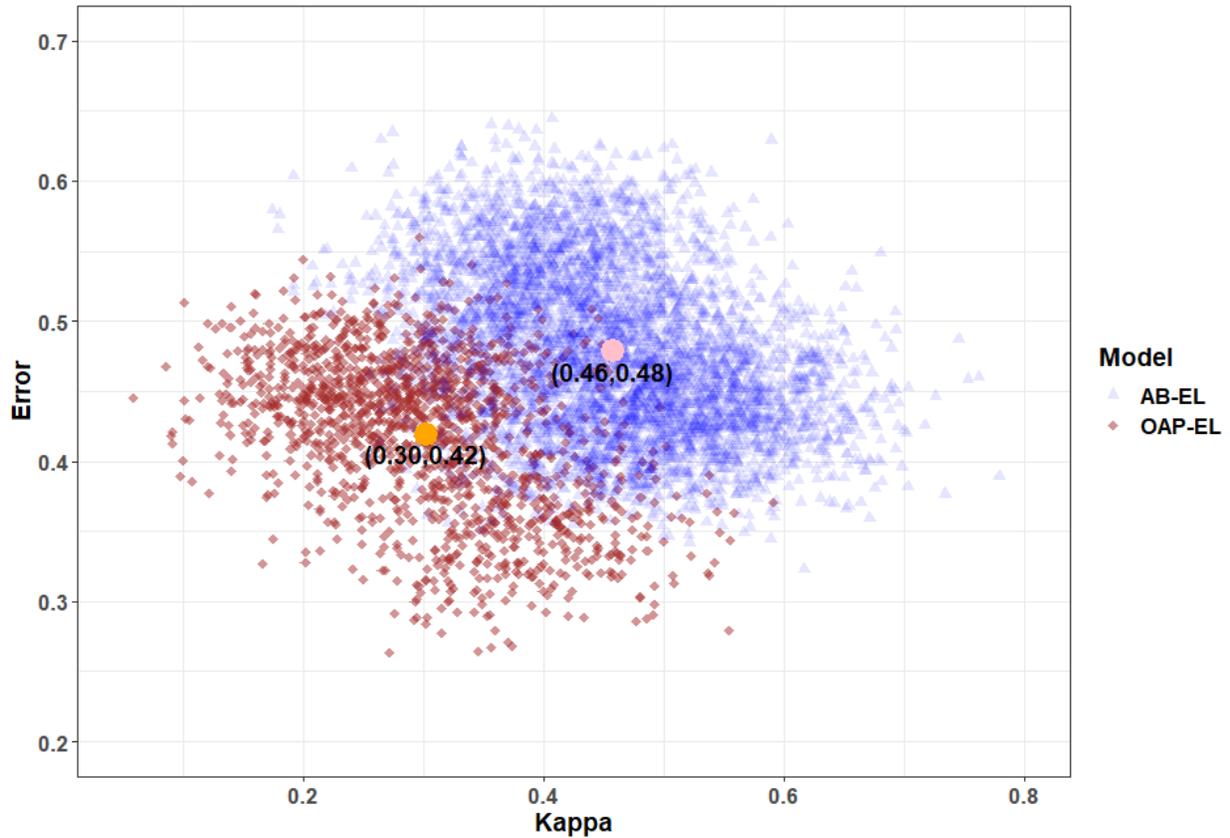

**Figure 14.** Kappa-error diagram for AB-EL and OAP-EL models. The x-coordinate of point represents the pairwise classifier diversity kappa measure, and y-coordinate is the averaged misclassification error of each pair of base-classifiers. AB-EL: Attribute Bagging Ensemble Learning; OAP-EL: Ontology-guided Attribute Partitioning Ensemble Learning. The highlight points indicate the mean of two groups.

### 4.3. Classifier Ensemble vs. Feature Ensemble

The current study proposes to integrate multiple classifiers, each of which is a *single-channel* classifier trained using a subset of features. In contrast to this "classifier ensemble" approach, we can also train a *multi-channel* classifier to integrate all the feature subsets ("feature ensemble"), like what we have proposed in our prior work [49]. We have demonstrated that OAP-EL model ("classifier ensemble") performs better than OAP-mNN model ("feature ensemble") in this particular application in both internal and external validations. Since the feature partitioning scheme was exactly the same for both models, the performance difference is likely because that the multi-channel models often require a relatively large dataset to reach a converged stable training loss. The ensemble learning model has far fewer parameters, reducing the potential overfitting issue.

### 4.4. Impacts of Imbalanced Dataset

Both CINEPS and COEPS cohorts had an imbalanced dataset, since very preterm infants who develop cognitive deficits were commonly fewer than the ones who develop normally. Such imbalanced datasets tend to impact on two aspects of our study: model training and model evaluation. For the aspect of model training, with an imbalanced dataset, machine learning models are prone to become majority class classifiers, i.e., they fail to learn the concepts of the minority class. There are a smaller number of very preterm infants in the high-risk group compared to those in the low-risk group with the ratio of 1:2 in the CINEPS dataset. To handle this issue, we applied a data synthesis approach SMOTE-ENN that simultaneously generated new synthetic samples for minority class and randomly removed sample in the majority class that may adversely impact model learning.

For the aspect of model evaluation, we provided PR curves and calculated PRAUC for both internal and external validation. In the future clinical practice, it is better to identify correctly as many high-risk very preterm infants as possible so that early intervention when interventions can exert the greatest impact on prevention. PR curves would be a proper evaluation method to use when the dataset has a substantial imbalanced sample ratio. PR curves and PRAUC are able to provide a better view for model performance on positive samples in the minority class, where ROC curves and AUC might not truly reflect the model performance. In the PR curves, since the random guess baseline changes according to positive: negative sample ratios, it is critical to compare our model to random guess baseline. In the internal validation, the positive: negative sample ratio of the CINEPS dataset is 1:2, resulting a PRAUC of 0.33 for the random guess baseline. (**Figure 7,8,11**) Our OAP-EL model had the best PRAUC of 0.56 among all competing models. Similarly, in the external validation, the PRAUC of random guess baseline was 0.14, while the developed OAP-EL model achieved a PRAUC of 0.42, which 3 times better than the random guess baseline. (**Figure 12**)

**4.5. Most Discriminative Brain Maturation and Geometric features**

Using the 2-level feature ranking method, we identified 15 top discriminative brain maturation and geometric features. The most predictive feature ranked by the OAP-EL model is the thickness of right insular cortex. The thickness of left insula region (ranked $13^{th}$) was also included within our feature list. Insulae are deeply buried regions that separate the frontal and parietal lobes from the temporal lobe. They are involved in a diverse array of brain functions, including perception, compassion, self-awareness, and cognitive function [59]. Insula thickness has been positively associated with non-planning impulsivity, a widely-used measure reflecting emotional development and decision-making [60]. Thus, it is not surprising that our model identified insula thickness as a discriminative feature that is predictive of cognitive deficits in very preterm infants. Several other cognition-related brain regions were also identified. For example, our OAP-EL model found that the sulcal depth of the anterior part of left medial and inferior temporal gyri was significantly predictive of cognitive deficits. Previous studies have demonstrated that the middle and inferior temporal gyri are associated with language and semantic memory processing, visual perception, and multimodal sensory integration [61-64]. Another highly discriminative brain region was the sulcal depth of the anterior part of left cingulate gyrus. The cingulate gyrus has been recognized to be highly involved in emotion formation and processing, learning, and memory [65-67]. Considering the important role of the frontal lobe and occipital lobe in learning, interestingly, only one region from each of these regions were selected by our model as top 15 discriminative features for predicting cognitive deficits. Nevertheless, because cognitive function is highly distributed across the brain, the selection of other brain regions and maturation features that are also involved in learning and cognition and learning suggests that our proposed OAP-EL model is able to learn meaningful geometric features instead of being overfitted by random noise.

### 4.6. Study Limitations

The current study includes certain limitations. First, ontology graph construction may vary between different studies. There is no universal method regarding how to utilize domain knowledge to construct an ontology graph. In addition, we only constructed an unweighted ontology graph using prior domain knowledge. How to effectively utilize domain knowledge to construct a weighted ontology graph could be a very interesting future direction. Second, we applied a spectral graph clustering algorithm to partition features into multiple non-overlapping subsets. Partitioning features into overlapping subsets has not been considered in the current study. Third, the external validation dataset (i.e., COEPS cohort) is a small dataset, where the high-risk group only contains 10 subjects. This resulted in the fact that predictions on a few samples may cause a large variation on model performance (e.g., one correctly predicted very preterm infant may increase the sensitivity of a model from 60% to 70%). Thus, even though our model outperformed peer models in both internal and external validation in the current study, a larger external cohort is necessary to validate the generalizability of our proposed model. Finally, our OAP approach is not applicable if all features partition into the same category.

## 5. Conclusion

We presented a novel OAP enhanced ensemble learning model integrating brain maturation and geometric features obtained at term-equivalent age for early prediction of cognitive deficits at 2 years corrected age in very preterm infants. The predictive performance of our novel ensemble model was significantly higher than models using traditional machine learning and peer ensemble learning. The proposed technique will facilitate ensemble learning in general, by helping augment the diversity among the base classifiers. In the future, we are also interested in developing ontology aided machine learning methods to better understand and depict both brain radiomics and connectomics features.

**Funding**: This work was supported by the National Institutes of Health [R01-EB029944, R01-EB030582, R01-NS094200 and R01-NS096037]; Academic and Research Committee (ARC) Awards of Cincinnati Children's Hospital Medical Center. The funders played no role in the design, analysis, or presentation of the findings.

**Acknowledgements:** We sincerely thank our collaborators from the Cincinnati Infant Neurodevelopment Early Prediction Study (CINEPS) Investigators: Principal Investigator: Nehal A. Parikh, DO, MS. Collaborators (in alphabetical order): Mekibib Altaye, PhD, Anita Arnsperger, RRT, Traci Beiersdorfer, RN BSN, Kaley Bridgewater, RT(MR) CNMT, Tanya Cahill, MD, Kim Cecil, PhD, Kent Dietrich, RT, Christen Distler, BSN RNC-NIC, Juanita Dudley, RN BSN, Brianne Georg, BS, Cathy Grisby, RN BSN CCRC, Lacey Haas, RT(MR) CNMT, Karen Harpster, PhD, OT/RL, Lili He, PhD, Scott K. Holland, PhD, V.S. Priyanka Illapani, MS, Kristin Kirker, CRC, Julia E. Kline, PhD, Beth M. Kline-Fath, MD, Hailong Li, PhD, Matt Lanier, RT(MR) RT(R), Stephanie L. Merhar, MD MS, Greg Muthig, BS, Brenda B. Poindexter, MD MS, David Russell, JD, Kari Tepe, BSN RNC-NIC, Leanne Tamm, PhD, Julia Thompson, RN BSN, Jean A. Tkach, PhD, Jinghua Wang, PhD, Brynne Williams, RT(MR) CNMT, Kelsey Wineland, RT(MR) CNMT, Sandra Wuertz, RN BSN CCRP, Donna Wuest, AS, Weihong Yuan, PhD. We sincerely thank Jennifer Notestine, RN, and Valerie Marburger, NNP for serving as our Nationwide Children's study coordinators and Mark Smith, MS, for serving as the study MR technologist. We are most grateful to the families that made this study possible.

**Declarations of interest:** None.

**Ethics Statement**

In accordance with The Code of Ethics of the World Medical Association, this study was approved by the Institutional Review Boards of the Cincinnati Children's Hospital Medical Center (CCHMC) and Nationwide Children's Hospital (NCH). Written parental informed consent was obtained for each subject.

**Data and Code Availability**: Requests to access the data sets used in this study should be directed to the corresponding author with a formal data sharing agreement and approval from the requesting researcher's local ethics committee. The source code of proposed model is publicly accessible on GitHub: (https://github.com/jiaolang771/aicad/tree/main/OAP_EL_Early_Prediction).

**Author Contributions**: Zhiyuan Li: Methodology, Software, Validation, Formal analysis, Visualization, Writing- Original Draft. Hailong Li: Conceptualization, Methodology, Software, Validation, Visualization, Writing- Original Draft. Adebayo Braimah: Data curation, Validation, Writing - Review & Editing. Jonathan R. Dillman: Validation, Writing - Review & Editing. Nehal A. Parikh: Conceptualization, Resources, Validation, Writing - Review & Editing, Funding acquisition. Lili He: Conceptualization, Methodology, Validation, Writing - Review & Editing, Funding acquisition.